\documentclass[11pt,a4paper]{article}
\usepackage{authblk}
\usepackage[hyperref]{emnlp-ijcnlp-2019}
\usepackage{times}
\usepackage{latexsym}
\usepackage{graphicx}
\usepackage{array,colortbl,multirow,multicol,booktabs,ctable}
\newcommand{\hdrule}{\midrule[\heavyrulewidth]}

\usepackage{url}

\aclfinalcopy 

\title{Bad Form: Comparing Context-Based and Form-Based\\ Few-Shot Learning in Distributional Semantic Models}

\author[$\spadesuit\heartsuit$]{\textbf{Jeroen Van Hautte}}
\author[$\spadesuit$]{\textbf{Guy Emerson}}
\author[$\spadesuit\diamondsuit\clubsuit$]{\textbf{Marek Rei}}
\affil[$\spadesuit$]{Dept. of Computer Science \& Technology, University of Cambridge, United Kingdom}
\affil[$\diamondsuit$]{The ALTA Institute, University of Cambridge, United Kingdom}
\affil[$\clubsuit$]{Dept. of Computing, Imperial College London, United Kingdom}
\affil[$\heartsuit$]{TechWolf, Belgium}
\affil[ ]{\small \texttt{jeroen@vanhautte.be, gete2@cam.ac.uk, marek.rei@cl.cam.ac.uk}}

\date{}

\begin{document}
\maketitle
\begin{abstract}
Word embeddings are an essential component in a wide range of natural language processing applications. However, distributional semantic models are known to struggle when only a small number of context sentences are available. Several methods have been proposed to obtain higher-quality vectors for these words, leveraging both this context information and sometimes the word forms themselves through a hybrid approach. We show that the current tasks do not suffice to evaluate models that use word-form information, as such models can easily leverage word forms in the training data that are related to word forms in the test data. We introduce 3 new tasks, allowing for a more balanced comparison between models. Furthermore, we show that hyperparameters that have largely been ignored in previous work can consistently improve the performance of both baseline and advanced models, achieving a new state of the art on 4 out of 6 tasks.
\end{abstract}

\section{Introduction}
Word embeddings have impacted almost every aspect of NLP, proving effective in a wide range of use cases. Often used in the form of a pre-trained model, these vectors provide easy to use representations of semantic meaning. However, distributional models are known to struggle with words for which training data is sparse, often resulting in low-quality vector representations \citep{huang2012improving, adams2017cross}. The default approach in this case has historically been to ignore these rare words, preferring an incomplete view over an incorrect one \citep{Mikolov:20130be}. Another option is to use the surface form of a word to obtain a vector, leveraging morphological characteristics \citep{luong2013better} or subword embeddings \citep{Bojanowski:20160be}. As neither of these approaches fully resolves the problem, more techniques have been proposed for few-shot learning in distributional models. Each of these aims to correctly position a new word vector inside an existing semantic space. The challenge for few-shot learning is to find a position that accurately reflects the meaning of the word, even if only a small number of usage examples is available.

Making systems better at handling rare words is an obvious practical goal of few-shot learning, as it could substantially improve systems working with technical language or dialects. However, few-shot learning is also interesting from a human language learning perspective: unlike current-day distributional models, humans excel at learning meaning from sparse data through a process called `fast mapping' \citep{trueswell2013propose, lake2017building}. Lessons learned from psychology might prove effective in machines, and novel few-shot learning techniques might provide insight into fast mapping in humans.

\begin{figure}
    \centering
    \includegraphics[width=0.37\textwidth]{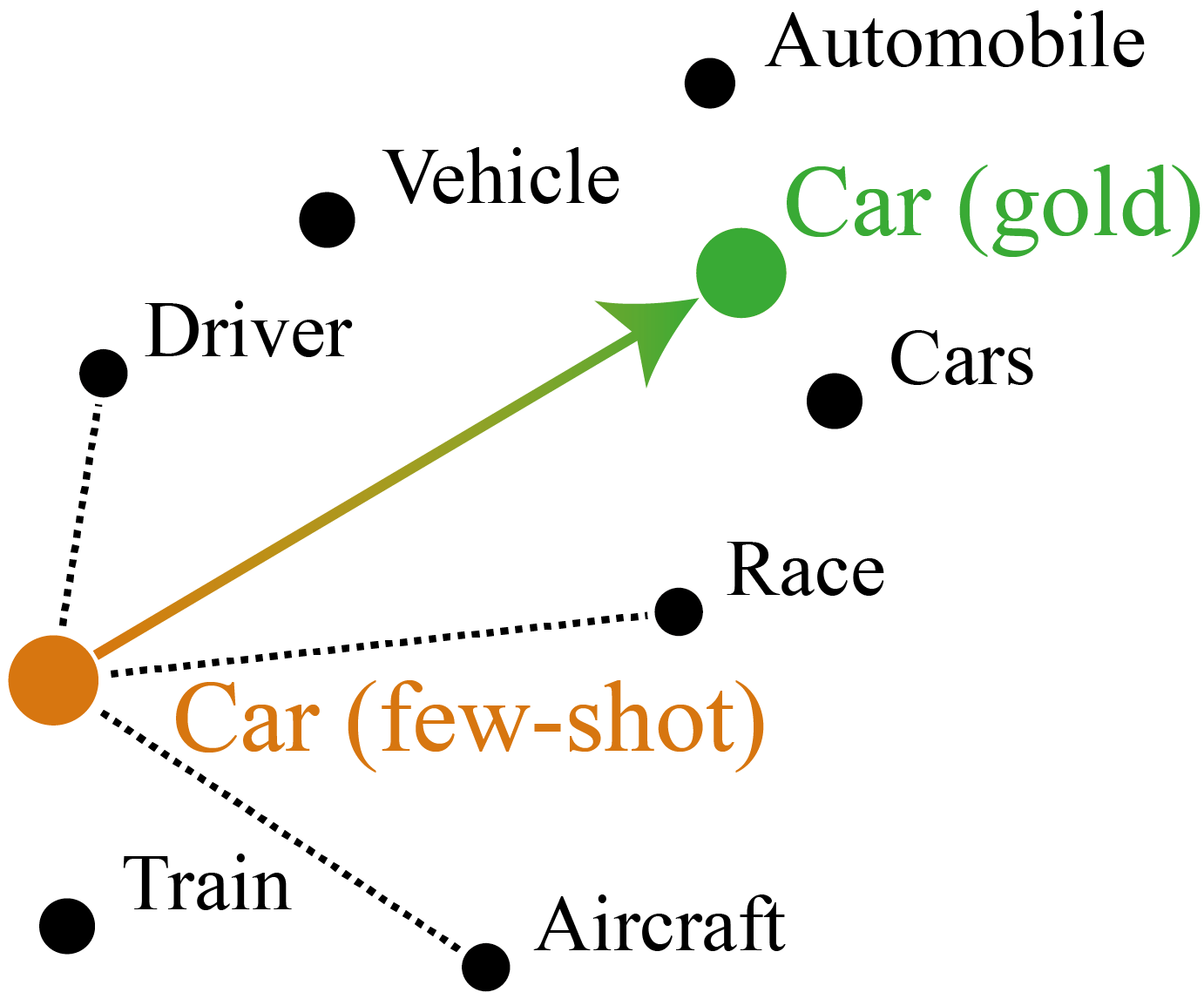}
    \caption{The DN task compares the few-shot vector to the gold vector (arrow), while the Chimera and CRW task compare system and human similarity to a selection of other words (dotted lines).}
    \label{fig:fewshot_eval}
\end{figure}


Three evaluation tasks have been proposed to evaluate few-shot learning methods: Definitional Nonce \citep{herbelot2017}, Chimera \citep{Lazaridou:20170be}, and Contextual Rare Words \citep{Khodak:20180be}, which we describe in Section~\ref{sec:back}. However, each of these tasks was designed for context-based few-shot learning, without considering hybrid methods, which also have access to word-form information. We show that the existing tasks do not suffice to fully assess the performance of hybrid models, with relatively simple, purely form-based methods dominating two out of three tasks. To provide a better overview and performance comparison, in Section~\ref{sec:eval} we introduce three new tasks based on these three datasets. In Sections~\ref{sec:novel}--\ref{sec:res}, we show that, just as hyperparameters are essential to good performance with standard distributional models \citep{levy2015improving}, the same is true for few-shot distributional models. With three straightforward modifications, we substantially improve the baseline scores, outperforming several advanced methods from previous work, as well as achieving a new state of the art on 4 out of 6 evaluation tasks.

\section{Background \& Related Work} \label{sec:back}
\subsection{Evaluation Tasks}
\label{sec:back:eval}
Three tasks have been used in most previous work to evaluate few-shot learning methods. The goal for each task is to obtain a high-quality vector for a word, given only a small set of sentences in which it appears. An existing semantic space is required,\footnote{In previous work, the model by \citet{herbelot2017} is often used.} in which a new vector needs to be placed. The embeddings in the existing semantic model are called background embeddings. A simple visualisation of the evaluation strategies is given in Figure~\ref{fig:fewshot_eval}.

\paragraph{Definitional Nonce} The Definitional Nonce (DN) task \citep{herbelot2017} provides a single definitional sentence for each test word. The test words are existing words, which have a high-quality gold vector, due to many occurrences in the training corpus. The aim for a few-shot learning algorithm is to infer a vector close to the gold vector. This is measured by ranking the background vectors by distance from the inferred vector, with the gold vector ideally placed at rank 1. The metrics used for this task are the Mean Reciprocal Rank (MRR) and median rank over the 300 test words. As the DN task uses definitional sentences as opposed to natural word use, we make use of the DN development set to optimise hyperparameters for this dataset separately.

\paragraph{Chimera} The Chimera dataset \citep{Lazaridou:20170be} consists of a series of novel words that are built as hybrids between two existing words. For each hybrid word, trials with 2, 4 and 6 context sentences are provided, with half of the sentences coming from each of the two source words. Each sentence was manually selected to be informative. Annotators were presented with the context sentences and asked to give similarity scores between the nonsense hybrid word and a range of other words. A few-shot learning algorithm is evaluated based on the rank correlation between the system's cosine similarity scores and the human similarity scores.

\paragraph{Contextual Rare Words} The Contextual Rare Words (CRW) dataset \citep{Khodak:20180be, luong2013better} consists of 255 context sentences selected randomly from Wikipedia for each of 455 existing words. Vectors are inferred for each word using 1, 2, 4, ..., 128 sentences. In similar fashion to the Chimera task, human similarity ratings to a selection of other words are compared to system ratings. For each number of context sentences, the Spearman rank correlation between the human and system similarities is reported. In this paper, we only report scores for up to 64 context sentences. We make use of the CRW development set introduced by \citet{Schick:20180be} to optimise model hyperparameters both for the CRW and Chimera task, as both of these have a similar setup. 

\subsection{Context-Based Few-Shot Learning}
\label{sec:back:context}
\paragraph{Word2Vec} While the Skip-Gram Word2Vec algorithm \citep{Mikolov:20130be} was used to generate the background embeddings provided by \citet{herbelot2017}, the method has also been applied as a few-shot learning method. This is done by loading the background embeddings and continuing training on the context sentences for each test word. This approach has been applied to each of the three tasks in previous work, with notably weak performance on the DN and Chimera datasets \citep{herbelot2017, Khodak:20180be, Schick:20180be}. However, to our knowledge, thorough hyperparameter optimisation for few-shot learning has not previously been attempted. 

\paragraph{Additive Model}
In similar fashion to \citet{herbelot2017}, we make use of a model that simply adds up all words in the context sentences for the test word. Stopwords\footnote{Based on the NLTK stopword list.} are dropped from this sum, as this has been found to consistently improve performance \citep{Khodak:20180be}.

\paragraph{Nonce2Vec} The Nonce2Vec algorithm heavily modifies several aspects of the standard Skip-Gram Word2Vec algorithm. This allows for a higher-risk initial learning approach, followed by a more cautious strategy as more data is presented \citep{herbelot2017}. 

\paragraph{Mem2Vec} The Mem2Vec algorithm uses a long-range memory over the whole corpus to find a vector corresponding to a small number of contexts \citep{sun2018}. 

\paragraph{A La Carte} The A La Carte model can be seen as an improved additive model: the addition is followed by a linear transformation, which is learned from the co-occurrence matrix of the corpus \citep{Khodak:20180be}. 

\subsection{Hybrid Few-Shot Learning}
\label{sec:back:hybrid}
In many words, part of the meaning can be deduced from the word form itself -- as such, models that can access and use this information can often perform better at few-shot learning.

\paragraph{FastText}
FastText \citep{Bojanowski:20160be} is an extension of Word2Vec: it is based on the same mechanisms, but adds in the use of character n-gram embeddings, as opposed to only modelling full words. The embedding for a word is calculated as the sum of its word embedding and the contained character n-gram embeddings. These are jointly optimised using the same approach as for Word2Vec. FastText is an interesting choice for few-shot learning due to its ability to generate vectors for out-of-vocabulary words: if a word is not contained in the vocabulary, a vector can be composed using only the character n-gram embeddings. 

\paragraph{Form-Context Model}
In similar fashion to FastText, the Form-Context Model \citep{Schick:20180be} combines both form and context information to infer a higher-quality vector. Two variants exist, both estimating the rare word vector $v_{(\mathbf{w}, \mathcal{C})}$ using:
\begin{equation}
v_{(\mathbf{w}, \mathcal{C})}=\alpha \cdot \hat{v}_{(\mathbf{w}, \mathcal{C})}^{{ context }}+(1-\alpha) \cdot v_{(\mathbf{w}, \mathcal{C})}^{{ form }}
\end{equation}
where $v_{(\mathbf{w}, \mathcal{C})}^{{ form }}$ is the surface form embedding and $\hat{v}_{(\mathbf{w}, \mathcal{C})}^{{ context }}$ is the context-based vector. The former is obtained through the subword approach from FastText\footnote{These subword embeddings are trained on top of an existing model, unlike those in FastText.}, while the latter vector is obtained through the A La Carte method. The two versions differ in their coefficient $\alpha$: in the single-parameter variant, $\alpha$ is a learned constant between $0$ and $1$, while in the gated model, it is a learned function of $v_{(\mathbf{w}, \mathcal{C})}^{ { form }}$ and $v_{(\mathbf{w}, \mathcal{C})}^{{ context }}$, allowing the model to adapt to different scenarios.

\section{Evaluation Setup} \label{sec:eval}
Several issues can be observed in the evaluation setup used in previous work. First of all, results on the Chimera task are inconsistent, showing almost no trends between different models. This can largely be attributed to the the size of the test set: only 110 chimera words are used. By using the CRW development set to optimise for both the CRW and Chimera task, we can include the training set as well, resulting in the `Full Chimera Task' with a total of 330 words.

For the CRW and DN tasks, the issues are not in the consistency of results, but rather in how to interpret the results. \citet{Schick:20180be} observe that, on the CRW task, their form-only model outperforms the full model, which uses both form and context. Whereas in context-based learning, each test word is new by definition, hybrid models typically have access to the vectors of different forms of the same lemma (such as \textit{wanderer} and \textit{wanderers}), meaning data for these words might not be sparse at all once related word forms are considered.

\begin{figure}
    \centering
    \includegraphics[width=0.5\textwidth]{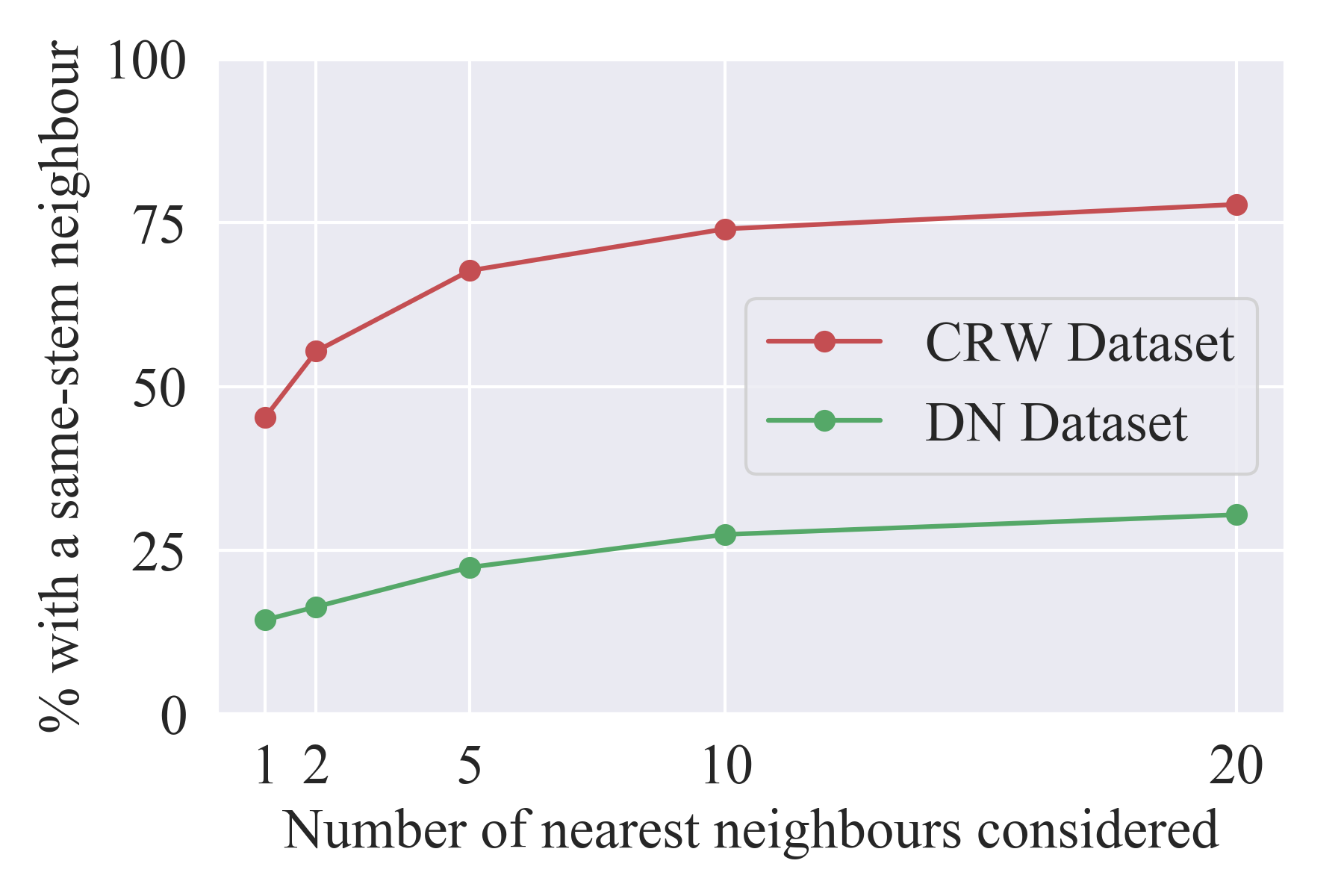}
    \caption{The proportion of words in the CRW and DN test sets that have a neighbour with the same stem, for different numbers of neighbours considered.}
    \label{fig:form_leakage}
\end{figure}

To assess the extent of the available information, we analyse the DN and CRW datasets, looking for words with the same stem\footnote{Determined using the NLTK Snowball stemmer.} as a test word. We ranked these words against the test word's nearest neighbours, with the results shown in Figure~\ref{fig:form_leakage}. For the CRW task, more than 50\% of test words have a word with the same stem among their 2 nearest neighbours, with this percentage increasing to more than 75\% when we look at the 20 nearest words. This indicates that there is a very high degree of information available to form-based methods that can leverage inflectional morphology. For the Definitional Nonce task, about 28\% of test words have a word with the same stem among their 20 nearest neighbours.

Based on these insights, we build a simple baseline model that estimates vectors by averaging all vectors for (non-test) words with the same stem. With an MRR of 0.5550 and a median rank of 2 on the DN task, this substantially outperforms the previous best scores of 0.1754 and 49 set by the Form-Context Model \citep{Schick:20180be}. On the CRW task, the model achieves a score of 0.32, well below the 0.49 score achieved by the form-only model from \citeauthor{Schick:20180be}. For both tasks, the top-performing model completely ignores the provided context sentences, implying that both of these tasks focus on new forms of known lemmas, rather than completely novel words.

Both scenarios are important use cases for few-shot distributional semantics, but to get a better view of the second scenario, we introduce the `Filtered CRW' and `Filtered DN' tasks, which have the same objectives as their \textit{non-filtered} counterparts, but for which the background embeddings are trained on a restricted corpus, filtering out any words with the same stem as one of the test words. This causes a removal of 2\% of the tokens inside the corpus, in similar fashion to how infrequent words\footnote{Based on a minimum count, for which 50 is used throughout this paper.} are typically dropped before a distributional model is trained. For the CRW dataset, the filtering removes the other word in 83 of the word pairs (for which human similarity scores are available), leaving the filtered version with 479 word pairs.

\section{Novel Methods} \label{sec:novel}
We now propose several novel methods, improving both baselines and advanced models through relatively simple modifications.

\subsection{Selective Word2Vec \& FastText}
\label{sec:novel:selective}
In previous work the default Word2Vec and FastText implementations are used. This means that not only do the vectors for test words change, but also those for context words. This creates a conflict of interest: to speed up learning of the new vector, a high learning rate might be desirable, but this same learning rate could also distort the background embeddings more heavily, decreasing vector quality. As such, we ensure that only vectors for test words are updated, removing the latter effect.\footnote{We also ensure that test words cannot be used as negative samples, so as to make sure that there is no influence between different samples.}

\subsection{Weighted Addition}
\label{sec:novel:weighted}
The additive model, A La Carte model and Form-Context model all make use of a simple, uniformly weighted sum\footnote{Followed by a transformation for ALC and FCM.} of all words around the test word. However, Word2Vec uses several techniques to focus on those words that are more likely to be meaningful. Below, we describe how these principles are included into each of these models by modifying the weights used in the addition.  

\subsubsection{Window Weights}
\begin{figure}
    \centering
    \includegraphics[width=0.45\textwidth]{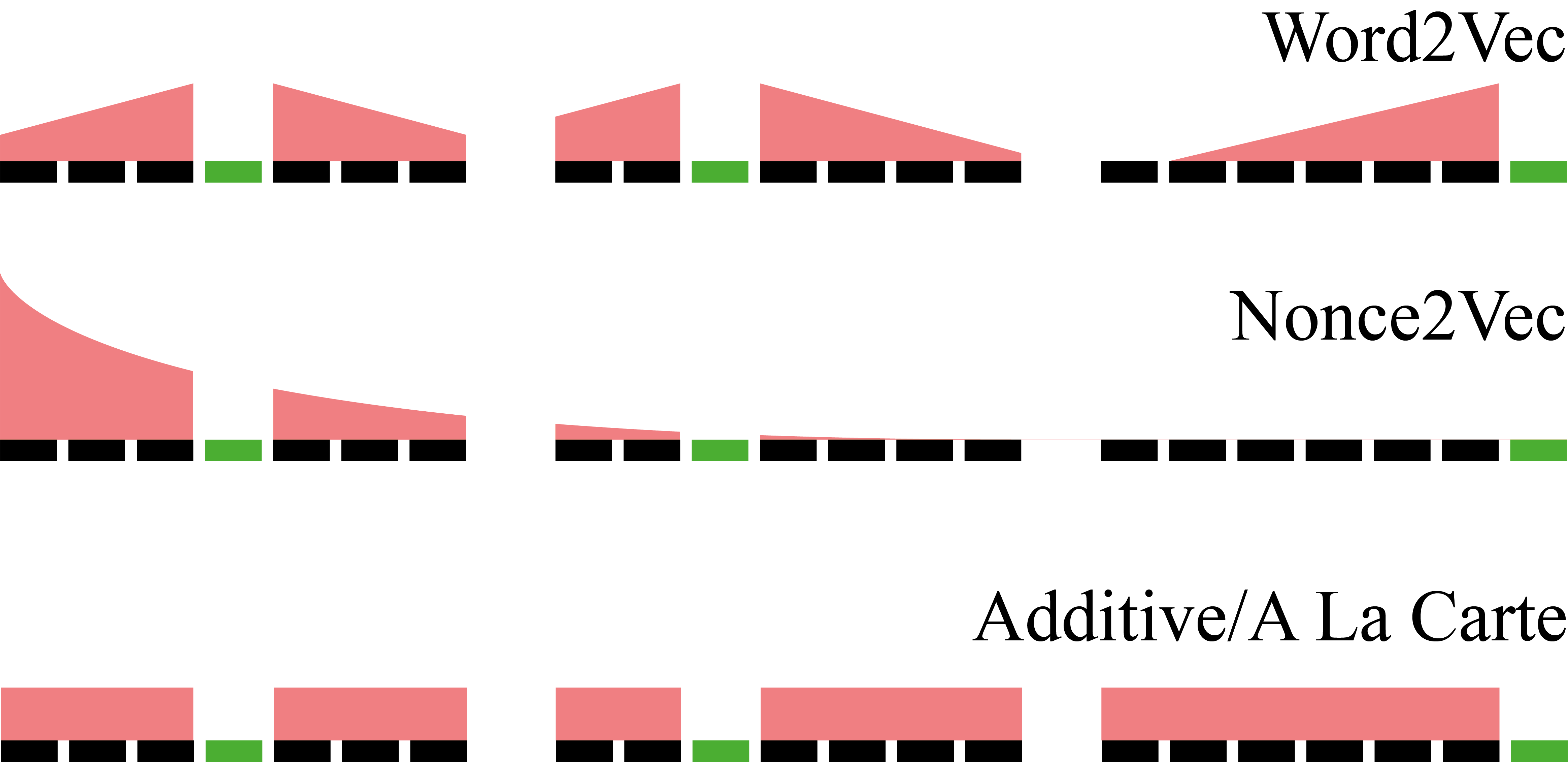}
    \caption{A visual representation of how different methods award importance to words, based on their position relative to the target word (green) and the order in which sentences are processed. The importance of a word can be interpreted as a combination of the selection probability and weight or learning rate used.}
    \label{fig:window_methods}
\end{figure}

The existing models have different strategies to handle the distance at which words co-occur, as shown in Figure~\ref{fig:window_methods}: while the additive and A La Carte model use each word in a sentence with the same weight, the exponential parameter decay of Nonce2Vec results in an emphasis on words before the test word. Word2Vec itself uses a sampled window size, meaning the importance of words decreases linearly with distance from the test word \citep{goldberg2014word2vec}. We adopt the same approach, as this has been shown to improve performance \citep{levy2015improving}. In Word2Vec, given a window size $n$, a context word $m$ tokens away from the target word has a probability of 
\begin{equation}
    P_{window}\left(m\right)={max}\left(\frac{n-m+1}{n}, 0\right)
\end{equation} 
to be selected as a positive sample. This probability can be seen as the expected weight of the contribution of each word to the final vector, which is how we apply it to the sum in each model.

\subsubsection{Negative Sampling \& Subsampling}
Word2Vec makes use of subsampling, as rare words typically carry more information than overly frequent words \citep{Mikolov:20130be, ramos2003using}. For a frequency threshold $t$ (typically $10^{-5}$), the probability to keep a word $w_i$ with frequency $f\left(w_{i}\right)$ in the training corpus is given by
\begin{equation}
    P_{subsample}\left(w_{i}\right)=1-\sqrt{\frac{t}{f\left(w_{i}\right)}}
\end{equation}
We again use this probability as a weight, multiplying each term with its subsampling probability.

Negative sampling has also been found to improve vector quality in Word2Vec \citep{goldberg2014word2vec}. The probability for a word $w_j$ to be selected as a negative sample is given by
\begin{equation}
    P_{negative}\left(w_{j}\right)=\frac{f\left(w_{j}\right)^{3 / 4}}{\sum_{k=0}^{n}\left(f\left(w_{k}\right)^{3 / 4}\right)}
\end{equation}
As such, the expected negative sample vector is 
\begin{equation}
    \hat{v}_{neg} = \sum_{w_i \in V}P_{negative}\left(w_{i}\right)\cdot v_{w_i}
\end{equation}
where $V$ is the vocabulary and $v_{w_i}$ is the vector for $w_i$. For a negative sampling rate $k$, each vector $v_{w_i}$ is now replaced with $v_{w_i} - k \cdot \hat{v}_{neg}$ before being added up, after which the subsample and window probabilities are applied.

\subsection{Neural ALC \& FCM}
\label{sec:novel:neural}
The A La Carte model uses a linear transformation, which is efficient, but also severely constrained \citep{Khodak:20180be}. We propose using a neural network with one hidden layer, allowing for a more flexible transformation. This same neural transformation can then be integrated into a new version of the Form-Context Model as well.

\section{Implementation Details}
\label{sec:implement}
We optimise the hyperparameters on the CRW and DN development sets, using the original evaluation setup. The same settings are used for the novel tasks. All model training described below is done using the Westbury Wikipedia Corpus \citep{shaoul2010westbury}, which was used in previous work \citep{Schick:20180be, Khodak:20180be}.

\subsection{Background Embeddings}
\label{sec:implement:background}
To make the experimental setup less complex, a single background model is used for all new tasks -- this can influence scores, but results for different models on the same task can still be compared. Filtering is applied to remove all words with the same stem as a test word, except for the DN words themselves, as these are used as gold vectors during evaluation.

To optimise the background embeddings (for both Word2Vec and FastText), we carry out a grid search for the learning rate (0.025, 0.05 or~0.1), dimension (150, 300 or~450), window size (5 or~10), negative sampling rate (5, 10 or~15) and number of epochs (5, 10 or~100). For FastText, the standard character n-gram length of 3 to~5 characters is used. Evaluating these models on the CRW development set \citep{Schick:20180be}, we find that for both algorithms, the default configuration\footnote{Learning rate~0.025, dimension 300, negative sampling~5, window size~5, and 5~epochs.}, while slightly outperformed by a higher-dimensional setup, is not significantly\footnote{At the 0.01 level, using a Monte Carlo permutation test.} worse than the respective top performer for each algorithm. Therefore, we opt for the default configuration in both cases.

\subsection{Previous Work}
\label{sec:implement:previous}
\subsubsection{Context-Based Few-Shot Learning}
\paragraph{Word2Vec}
We conduct the same grid search as for the background embeddings (aside from the dimension, which is fixed to 300). For the CRW dataset, we again find the same parameters. For the DN task, a higher number of epochs (100) and window size (10) substantially improve upon the default setup. A possible explanation for the big difference in hyperparameters can be found in the negative sampling mechanism: as only a single context sentence is used in the DN task, a small number of negative samples is used per epoch. This makes the effect of negative sampling on the returned vector more variable, while the gold vector has had sufficient (positive and negative) samples to converge to the expected value. By raising the window size and the number of epochs, the number of negative samples is also increased, allowing for the same to happen even with a single input sentence. On the CRW task, there is no similar tendency, indicating that this `expected effect of negative sampling' is important for recreating a vector exactly, but less so for the quality of the vector as compared to human judgement. 

\paragraph{Nonce2Vec} For performance reasons, we implement the Nonce2Vec algorithm ourselves. We use the same parameters reported by \citet{herbelot2017} and confirm that results are comparable to the original.

\paragraph{Mem2Vec} As no code was published for Mem2Vec, we have not evaluated the algorithm ourselves. For completeness, we report the results available for the DN and CRW tasks.

\paragraph{A La Carte} The code provided with the original paper allows us to easily generate the transformation matrix needed for this model.

\begin{table*}[ht]
\small
    \centering
    \setlength\tabcolsep{3pt} 
    \begin{tabular}{cl|cc|cc|ccc|ccc}
    \toprule
    & & \multicolumn{2}{c|}{\textbf{Definitional Nonce}} & \multicolumn{2}{c|}{\textbf{Filtered DN}} & \multicolumn{3}{c|}{\textbf{Chimera}} & \multicolumn{3}{c}{\textbf{Full Chimera}}\\
    &\textbf{Method} & MRR & Median & MRR & Median & L2 & L4 & L6 & L2 & L4 & L6\\
    \midrule
    \multirow{11}{*}{\rotatebox{90}{\textbf{CONTEXT BASED}}}&Word2Vec & 0.0007 & \hphantom{0}5253 & 0.0110 & \hphantom{0}3546& 0.299 & 0.332 & 0.404 & 0.265 & 0.355 & 0.363\\
    &Additive & 0.0332 & \hphantom{00}870 & 0.0377 & \hphantom{00}678 & 0.358 & \textbf{0.387} & 0.420 & 0.320 & 0.366 & 0.388\\
    &Nonce2Vec & 0.0415 & \hphantom{00}708 & 0.0557 & \hphantom{00}583 & 0.328 & 0.378 & 0.401 & 0.300 & 0.356 & 0.369\\
    &A La Carte & \textbf{0.0706} & \hphantom{00}\textbf{165} & \textbf{0.0697} & \hphantom{00}\textbf{155} & \textbf{0.363} & 0.384 & 0.394 & 0.304 & 0.355 & 0.377\\
    &Mem2Vec & 0.0542 & \hphantom{00}512 & - & - & 0.330 & 0.372 & 0.390 & - & - & -\\
    \cmidrule{2-12}
    &Selective Word2Vec & 0.0183 & \hphantom{0}1710 & 0.0255 & \hphantom{0}1570 & 0.301 & 0.323 & 0.410 & 0.270 & 0.343 & 0.365\\
    \cmidrule{2-12}
    &Additive + Window & 0.0364 & \hphantom{00}937 & 0.0320 & \hphantom{00}646 & 0.359 & 0.370 & \textbf{0.433} & \textbf{0.327} & \textbf{0.369} & \textbf{0.391}\\
    &Additive + Window/Sub/Neg & 0.0523 & \hphantom{00}267 & 0.0400 & \hphantom{00}418 & 0.360 & 0.355 & 0.422 & 0.314 & 0.356 & 0.388\\
    \cmidrule{2-12}
    &A La Carte + Window & 0.0426 & \hphantom{00}637 & 0.0321 & \hphantom{00}591 & 0.292 & 0.376 & 0.390 & 0.288 & 0.348 & 0.372\\
    &A La Carte + Window/Sub/Neg & 0.0327 & \hphantom{0}2274 & 0.0323 & \hphantom{00}510 & 0.261 & 0.334 & 0.375 & 0.294 & 0.345 & 0.365\\
    &Neural A La Carte + Window & 0.0472 & \hphantom{00}931 & 0.0334 & \hphantom{0}1114 & 0.325 & 0.374 & 0.401 & 0.306 & 0.367 & 0.386\\

    \hdrule
    \multirow{4}{*}{\rotatebox{90}{\parbox{1.6cm}{\centering\textbf{FORM + \\ HYBRID}}}}& FastText & - & - & - & - & - & - & - & 0.129 & 0.165 & 0.202\\
    & Form-Context & 0.1561 & \hphantom{000}64 & \textbf{0.0992} & \hphantom{000}\textbf{99} & 0.325 & \textbf{0.367} & 0.359 & \textbf{0.313} & 0.339 & 0.333\\
    \cmidrule{2-12}
    & Stem-Based & \textbf{0.5550} & \hphantom{0000}\textbf{2} & - & - &-&-&-&-&-&-\\
    & Selective FastText & - & - & - & - & - & - & - & 0.060 & 0.087 & 0.120 \\
    & Neural FCM & 0.1219 & \hphantom{00}183 & 0.0735 & \hphantom{00}241 & \textbf{0.327} & 0.361 & \textbf{0.382} & 0.304 & \textbf{0.351} & \textbf{0.360}\\
    \bottomrule
    \end{tabular}
    \caption{Results for DN and Chimera tasks. The best result per category (context-based or hybrid) in every column is marked in bold, while setups that were not evaluated have been filled with a dash. Purely form-based methods are not evaluated on the Chimera tasks. The stem-based model is not compatible with the filtered tasks.}
    \label{tab:dn_chimera_results}
\end{table*}

\subsubsection{Hybrid Few-Shot Learning}

\paragraph{FastText}
For the CRW task, we again find the default settings to be optimal. We only evaluate FastText on the Filtered CRW and Full Chimera tasks, so as to avoid problems with model dependence (DN task) and lexical information leakage.

\paragraph{Form-Context Model}
We make use of the gated model, training it just like \citet{Schick:20180be}. The same character n-gram lengths are used as for FastText.

\subsection{Novel Methods}
\paragraph{Selective Word2Vec \& FastText} Based on Gensim's Skip-Gram implementation, we create a selective version of Word2Vec and FastText. All parameters are found to be the same as for the non-selective versions, except for selective Word2Vec on the DN task, where a higher-risk learning rate of 0.1 is now optimal.

\paragraph{Window Weights} To add window weights to the addition-based models, we evaluate both the additive and A La Carte models with a window size of $2, 5, 10, 15$ and $20$, finding $10$ to be optimal across the board.

\paragraph{Subsampling \& Negative Sampling} For the subsampling mechanism, the frequency threshold $t = 10^{-5}$ is used, as recommended by \citet{Mikolov:20130be}. For the negative sampling mechanism, rates of $1,2,5$ and $10$ negative samples per positive sample are considered, with $2$ being optimal for both the CRW and DN development set.

\paragraph{Neural ALC \& FCM}
We use a simple architecture with one hidden layer. This hidden layer has 1000 neurons (out of 100, 200, 500, 1000 and 2000) and has a ReLU activation. The output layer has no non-linearity. The network is optimised with the Adam optimiser \citep{kingma2015adam} and the mean square error loss function. The model is trained with the same samples as the original A La Carte model. The same window weights are used as before.

\begin{table*}[ht]
\small
    \centering
    \setlength\tabcolsep{2.5pt} 
    \begin{tabular}{cl||c|ccccccc||c|ccccccc}
    \toprule
    & & \multicolumn{8}{c||}{\textbf{Contextual Rare Words}} & \multicolumn{8}{c}{\textbf{Filtered CRW}}\\

    & \multicolumn{1}{c||}{}& \multicolumn{8}{c||}{{Number of Context Sentences}} & \multicolumn{8}{c}{{Number of Context Sentences}} \\
    & \textbf{Method} & 0 & 1 & 2 & 4 & 8 & 16 & 32 & 64 & 0 & 1 & 2 & 4 & 8 & 16 & 32 & 64\\
    \midrule
    \multirow{12}{*}{\rotatebox{90}{\textbf{CONTEXT BASED}}}&Word2Vec & - & 0.15 & 0.19 & 0.24 & 0.30 & 0.35 & 0.38 & 0.41 & - & 0.14 & 0.18 & 0.23 & 0.29 & 0.33 & 0.37 & 0.39\\
    &Additive & - & 0.12 & 0.14 & 0.16 & 0.17 & 0.18 & 0.19 & 0.19 & - & 0.11 & 0.13 & 0.15 & 0.17 & 0.18 & 0.18 & 0.18 \\
    &Nonce2Vec & - & 0.12 & 0.15 & 0.17 & 0.17 & 0.16 & 0.15 & 0.14 & - & 0.12 & 0.15 & 0.17 & 0.18 & 0.17 & 0.16 & 0.15\\
    &A La Carte & - & 0.19 & 0.24 & 0.29 & 0.34 & 0.38 & 0.40 & 0.40 & - & 0.20 & 0.25 & 0.29 & 0.34 & 0.37 & 0.39 & 0.40\\
    \cmidrule{2-18}
    &Selective Word2Vec & - & 0.19 & 0.23 & 0.27 & 0.32 & 0.36 & 0.38 & 0.40 & - & 0.17 & 0.22 & 0.26 & 0.31 & 0.34 & 0.37 & 0.39\\
    \cmidrule{2-18}
    &Additive + Window & - & 0.13 & 0.16 & 0.18 & 0.20 & 0.20 & 0.21 & 0.21 & - & 0.14 & 0.17 & 0.20 & 0.22 & 0.23 & 0.24 & 0.24\\
    &Additive + Window/Sub/Neg & - & 0.16 & 0.20 & 0.24 & 0.27 & 0.29 & 0.30 & 0.30 & - & 0.17 & 0.21 & 0.24 & 0.27 & 0.29 & 0.30 & 0.31\\
    \cmidrule{2-18}
    &ALC + Window & - & 0.21 & 0.26 & 0.32 & 0.36 & 0.40 & 0.41 & 0.42 & - & 0.22 & 0.27 & 0.32 & 0.36 & \textbf{0.39} & 0.40 & 0.41\\
    &ALC + Window/Sub/Neg & - & 0.21 & 0.26 & 0.31 & 0.34 & 0.35 & 0.36 & 0.36 & - & \textbf{0.24} & \textbf{0.29} & \textbf{0.33} & \textbf{0.37} & \textbf{0.39} & \textbf{0.41} & \textbf{0.42}\\
    &Neural ALC + Window & - & \textbf{0.22} & \textbf{0.27} & \textbf{0.33} & \textbf{0.37} & \textbf{0.41} & \textbf{0.43} & \textbf{0.44} & - & 0.23 & 0.28 & \textbf{0.33} & \textbf{0.37} & \textbf{0.39} & \textbf{0.41} & \textbf{0.42} \\
    \hdrule
    \multirow{6}{*}{\rotatebox{90}{\parbox{2.5cm}{\centering\textbf{FORM + \\ HYBRID}}}} & FastText & -&-&-&-&-&-&-&- & \textbf{0.36} & 0.31 & 0.32 & 0.32 & 0.32 & 0.33 & 0.33 & 0.34\\
    & Form-Context & \textbf{0.49} & 0.42 & \textbf{0.45} & \textbf{0.46} & \textbf{0.47} & \textbf{0.47} & 0.47 & 0.47 & \textbf{0.36} & 0.32 & 0.35 & 0.37 & \textbf{0.39} & {0.39} & 0.39 & 0.40\\
    \cmidrule{2-18}
    & Stem-Based & 0.32 &-&-&-&-&-&-&-&-&-&-&-&-&-&-&-\\
    & Selective FastText & -&-&-&-&-&-&-&- & \textbf{0.36} & \textbf{0.36} & \textbf{0.36} & 0.37 & 0.37 & 0.38 & {0.40} & {0.42}\\
    &Neural FCM & \textbf{0.49} & \textbf{0.43} & \textbf{0.45} &  \textbf{0.46} & \textbf{0.47} & \textbf{0.47} & \textbf{0.48} & \textbf{0.48} & \textbf{0.36} & 0.33 & \textbf{0.36} & \textbf{0.38} & \textbf{0.39} & \textbf{0.41} & \textbf{0.42} & \textbf{0.43}\\
    \bottomrule
    \end{tabular}
    \caption{Results on both CRW tasks. The best result per category in every column is marked in bold. The form-based and hybrid categories are shown together, as a hybrid model using 0 context sentences is effectively form-based. The stem-based model is not compatible with the filtered tasks.}
    \label{tab:crw_results}
\end{table*}

\section{Results} \label{sec:res}
We now discuss the results for each dataset, with the emphasis on trends in how models adapt to different circumstances. A summary is provided in Section~\ref{sec:conc}.

\subsection{Definitional Nonce \& Filtered DN}
Results for the DN and Filtered DN task are shown in Table~\ref{tab:dn_chimera_results}. The best context-based model on both tasks is the A La Carte model, which significantly\footnote{Significance testing is applied to the MRR metric.} outperforms all other context-based models. While the Form-Context model performs significantly better than A La Carte on both tasks, the original DN task is completely dominated by the stem-based model. This shows that using known related word forms is an extremely effective approach for estimating a new embedding. The removal of these related words from the training data heavily impacts the scores for all form-based and hybrid methods, but the Form-Context model still manages to perform strongly on the Filtered DN task, showing that the model is robust to varying amounts of information in both the form and context.

Selective Word2Vec and the additive models show that the baseline scores reported by \citeauthor{herbelot2017} left much room for improvement, but both normal and selective Word2Vec are still among the worst models evaluated. The effect of weights used in the addition-based models is not consistently positive, which might be explained by the fact that these principles are meant for natural word usage (not definitions, as in the DN dataset).

\subsection{Chimera \& Full Chimera}
The results for the original and Full Chimera tasks are provided in Table~\ref{tab:dn_chimera_results}. On the original task, there are no clear performance trends, presumably caused by the small size of the test set. In that respect, the Full Chimera task is much more useful, allowing for a better comparison between models. The additive model with a window achieves the best score on all trials for the Full Chimera task, as well as one for the original (L6, with strong performance L2 and L4 as well). The most advanced additive model (window, subsampling and negative sampling) and the nonlinear A La Carte model also perform very strongly on the Full Chimera task.

There is a clear divide between the context-based and hybrid models, with the latter being outperformed by almost all of the former. This is caused by the nonsensical word forms used for chimeras: the form information is now misinformation. Looking at the large performance difference between both Form-Context models and the FastText-based models, the advantage of the Form-Context architecture becomes clear: the adaptive weighting between form and context provides much better flexibility. FastText, on the other hand, has a fixed strategy, meaning it cannot disregard the useless form information. The original FastText algorithm outperforms selective FastText, as the subword embeddings are able to overfit on the provided context sentences. 

\subsection{CRW Tasks}
The results for the CRW tasks are provided in Table~\ref{tab:crw_results}. On the original CRW task, hybrid methods dominate, with the FCM outperforming all other results significantly even with no context sentences. This again shows how much lexical information is available.

On the Filtered CRW task, form-based scores are much lower. However, as shown by the two FastText models and the two Form-Context models, using form information can still provide a clear advantage here by augmenting sparse context-based information. In this situation, the fixed strategy used by FastText allows the selective FastText algorithm to be among the top models on the Filtered CRW task, while the original FastText algorithm suffers from overfitting. The best model overall is the Neural FCM, achieving the top result on all but one trial.

Among the context-based methods, all A La Carte models perform strongly, just like the selective Word2Vec algorithm. On the Filtered CRW task, the integration of the principles behind Word2Vec consistently improves performance, both for the additive and A La Carte models, showing that these are particularly effective when working with unfiltered, natural usage examples for new words. In the original CRW task however, these techniques cause a performance drop, most likely caused by the presence of the original words in the model. The Neural ALC model is the best context-based model on both tasks: the extra freedom allowed by the neural network allows this model to adapt better to different situations. Interestingly, performance for Nonce2Vec decreases when more than 16 context sentences are used. This is seems to be caused by the imbalance in the importance of training data (Figure~\ref{fig:window_methods}).

\section{Conclusion} \label{sec:conc}
Different situations and goals in few-shot learning have different optimal solutions. The difference between learning from natural language usage and definitions is especially apparent: only the original A La Carte method performs well for both types, while other models that do very well on the latter typically trail on the former. The principles behind Word2Vec work well in other models when using unfiltered, natural usage examples, but are less consistent when the sentences are filtered (Chimera dataset) or of a different type (DN dataset). The available word-form information is a double-edged sword: while real-world scenarios will often allow for the use of such information, a completely novel word form can cause a decrease in performance. With a combination of existing and novel evaluation tasks, we have been able to compare and explain model performance between context-based and hybrid methods in different scenarios.

The success of the newly proposed baseline methods shows that within specific use cases, a simple approach can suffice to achieve very strong performance. More complex methods, such as Nonce2Vec and Mem2Vec, are even outperformed across the board by these new baselines. However, simple methods typically struggle to generalise to multiple sub-tasks. The main benefit of more complex methods is that they are more flexible, at the price of overhead and a risk of overfitting. For both context-based and hybrid few-shot learning, we have achieved a new state of the art on 4 out of the 6 evaluation tasks used, showing that a careful, optimised approach can be the key to success in few-shot learning. Future work could explore other distributional models, such as dependency embeddings \citep{levy2014dependency,czarnowska2019dependency}, but it is clear from our results that careful optimisation will be required to adapt other models to the few-shot setting.

\section*{Acknowledgements}
We thank the anonymous reviewers for their valuable feedback. Guy Emerson is supported by a Research Fellowship at Gonville \& Caius College, Cambridge.

\bibliography{emnlp-ijcnlp-2019}
\bibliographystyle{acl_natbib}

\end{document}